\documentclass[10pt,twocolumn,letterpaper]{article}
\usepackage[accsupp]{axessibility}
\usepackage{iccv}
\usepackage{times}
\usepackage{epsfig}
\usepackage{graphicx}
\usepackage{csquotes}
\usepackage{verbatim}
\usepackage{color,soul}
\usepackage{epsfig}
\usepackage{amsmath}
\usepackage{amssymb}
\usepackage{datetime2}
\usepackage{bm}
\usepackage{microtype}      
\usepackage{dcolumn}        
\usepackage{url}            
\usepackage{booktabs}       
\usepackage{amsfonts}       
\usepackage{nicefrac}       
\usepackage{enumitem}
\usepackage{float}
\usepackage{stfloats}
\usepackage{algorithm}
\usepackage[noend]{algpseudocode}
\usepackage[numbers,sort,compress]{natbib}
\usepackage{amsthm}
\usepackage{subfiles}
\usepackage{adjustbox}
\usepackage{subcaption}
\usepackage{makecell}
\usepackage[normalem]{ulem}
\usepackage{caption}
\usepackage{nopageno}
\usepackage[flushmargin,para]{footmisc}
\usepackage[table]{xcolor}
\usepackage{pifont}
\usepackage{array}
\usepackage{arydshln} 
\usepackage{tabularx}
\usepackage{booktabs}

\usepackage{multirow}
\newcommand{\PreserveBackslash}[1]{\let\temp=\\#1\let\\=\temp}
\newcolumntype{C}[1]{>{\PreserveBackslash\centering}p{#1}}
\newcolumntype{R}[1]{>{\PreserveBackslash\raggedleft}p{#1}}
\newcolumntype{L}[1]{>{\PreserveBackslash\raggedright}p{#1}}
\newcommand{\cmark}{\ding{51}}%
\newcommand{\xmark}{\ding{55}}%

\newcommand{\methodname}{JOTR\xspace}
\definecolor{lightgray}{gray}{0.97}
\definecolor{lightblue}{rgb}{0.93,0.95,1.0}

\usepackage[pagebackref=true,breaklinks=true,letterpaper=true,colorlinks,bookmarks=false]{hyperref}

\usepackage[capitalize]{cleveref}
\crefname{section}{Sec.}{Secs.}
\Crefname{section}{Section}{Sections}
\Crefname{table}{Table}{Tables}
\crefname{table}{Tab.}{Tabs.}

\iccvfinalcopy 

\def\iccvPaperID{2019} 

\begin{document}
\title{JOTR: 3D Joint Contrastive Learning with Transformers for \\ Occluded Human Mesh Recovery}
\author{
		Jiahao Li$^{1,2}$\footnotemark[2], Zongxin Yang$^{1}$, Xiaohan Wang$^1$, Jianxin Ma$^2$, Chang Zhou$^2$, Yi Yang$^{1}$\footnotemark[3] \\
		$^1$ ReLER, CCAI, Zhejiang University \space \space \space $^2$ DAMO Academy, Alibaba Group \\
	}
\twocolumn[{%
		\renewcommand\twocolumn[1][]{#1}%
		\maketitle
		\begin{center}
			\newcommand{\teaserwidth}{\textwidth}
			\vspace{-6mm}
			\centerline{
				\includegraphics[width=0.99\teaserwidth,clip]{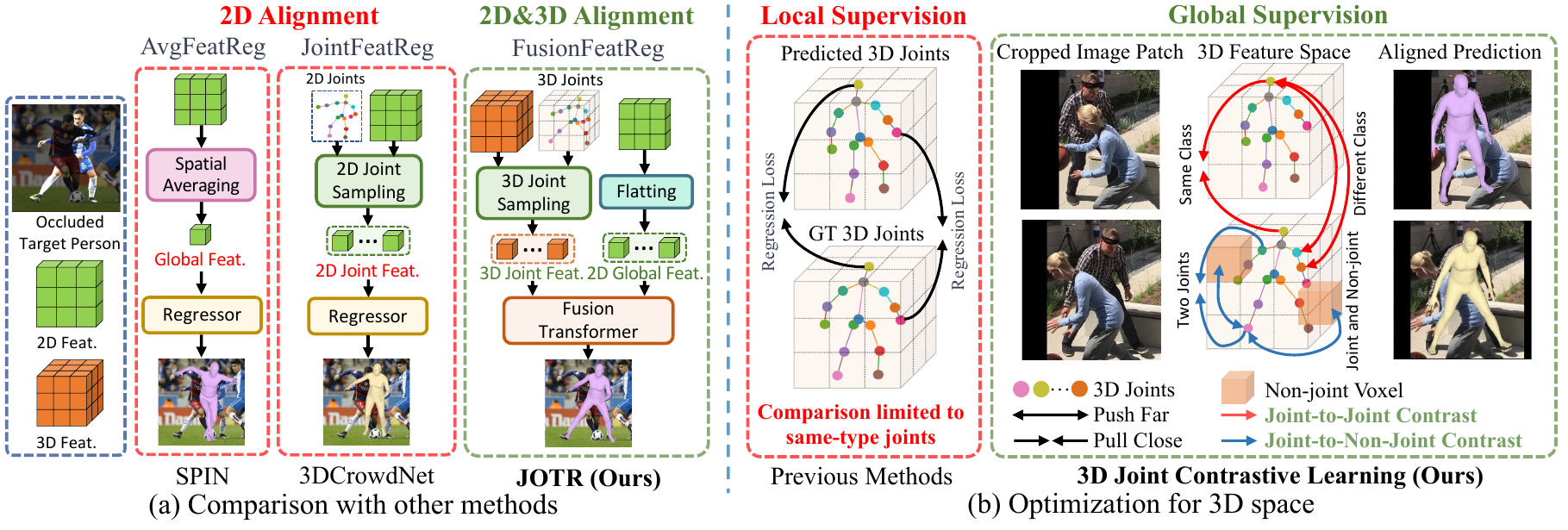}
			}
			\vspace{-0.1in}
			\captionof{figure}{
                (a) Current methods for human mesh recovery could be classified into two categories: AvgFeatReg~\cite{SPIN,HMR} and JointFeatReg~\cite{ROMP,zhang2021pymaf,3dCrwodNet}, which focus on improving 2D alignment by using average 2D features (feat.) for regression and employing sampled 2D joint feature for regression, respectively.
                In contrast, our proposed novel techniques (\ie, FusionFeatReg), fuse 2D and 3D features for regression to enhance both 2D and 3D alignment. (b) Moreover, to provide global supervision for the entire 3D space, we introduce a 3D joint contrastive learning method, which stands in contrast to previous approaches that solely apply 3D joints as local supervision.
			}
			\vspace{-4mm}
			\label{fig:fig_1}
		\end{center}%
}]



\renewcommand{\thefootnote}{\fnsymbol{footnote}}
\footnotetext[2]{Jiahao Li worked on this at his Alibaba internship.\space \space \space\space \space \space\space \space \space\space \space \space\space \space \space\space \space \space}
\footnotetext[3]{Yi Yang is the corresponding author.}

\begin{abstract}
    \vspace{-5mm}
In this study, we focus on the problem of 3D human mesh recovery from a single image under obscured conditions.
Most state-of-the-art methods aim to improve 2D alignment technologies, such as spatial averaging and 2D joint sampling.
However, they tend to neglect the crucial aspect of 3D alignment by improving 3D representations.
Furthermore, recent methods struggle to separate the target human from occlusion or background in crowded scenes as they optimize the 3D space of target human with 3D joint coordinates as local supervision.
To address these issues, a desirable method would involve a framework for fusing 2D and 3D features and a strategy for optimizing the 3D space globally. Therefore, this paper presents 3D JOint contrastive learning with TRansformers (\textbf{JOTR}) framework for handling occluded 3D human mesh recovery. Our method includes an encoder-decoder transformer architecture to fuse 2D and 3D representations for achieving 2D$\&$3D aligned results in a coarse-to-fine manner and a novel 3D joint contrastive learning approach for adding explicitly global supervision for the 3D feature space. The contrastive learning approach includes two contrastive losses: joint-to-joint contrast for enhancing the similarity of semantically similar voxels (i.e., human joints), and joint-to-non-joint contrast for ensuring discrimination from others (e.g., occlusions and background). Qualitative and quantitative analyses demonstrate that our method outperforms state-of-the-art competitors on both occlusion-specific and standard benchmarks, significantly improving the reconstruction of occluded humans. Code is available at \url{https://github.com/xljh0520/JOTR}.
\end{abstract} 

\vspace{-6mm}
\section{Introduction}
\label{sec:introduction}

The estimation of 3D human meshes from single RGB images is an active area of research in computer vision with a broad range of applications in robotics, AR/VR, and human behavior analysis.
In contrast to estimating the pose of general objects~\cite{yang2021dsc}, human mesh recovery is more challenging due to the complex and deformable structure of the human body. Nevertheless, enhancing human-centric tasks can be achieved by combining visual features and prior knowledge about human anatomy through constructing multi-knowledge representations~\cite{mkr}.
Generally, the human mesh recovery task takes a single image as input and regresses human model parameters such as SMPL~\cite{SMPL} as output.

Driven by deep neural networks, this task has achieved rapid progress~\cite{HMR,SPIN,3dCrwodNet,BEV,CVPE_2020_OOH,zhang2021pymaf,ROMP,PARE,OCHMR,jiang2020multiperson,metro_lin,GraphCMR,mesh_graphormer}. 
Recent studies have focused on regressing accurate human meshes despite occlusions.
To achieve this, most of them employ 2D prior knowledge (\eg, UV maps~\cite{CVPE_2020_OOH}, part segmentation masks~\cite{PARE} and 2D human key points~\cite{OCHMR}) to focus the model on visible human body parts for enhancing the 2D alignment of the predicted mesh. 
Additionally, some methods~\cite{3dCrwodNet,BEV} introduce 3D representations to locate 3D joints and extract 2D features from the corresponding regions of the 2D image.

Even though the above methods have achieved significant progress in occluded human mesh recovery, they still remain constrained to these two aspects: the pursuit of \textbf{2D alignment} and \textbf{local supervision} for 3D joints.
(\romannumeral1) As shown in~\cref{fig:fig_1}{\color{red}a}, the above methods employing 2D prior knowledge mainly focus on \textbf{2D alignment} technologies, including spatial averaging and 2D joint sampling.
However, in crowded or occluded scenarios, solely focusing on 2D alignment may acquire ambiguous features for the entire mesh due to the lack of estimation of hidden parts.
Accordingly, the invisible human body parts would be aligned based on prior knowledge of the standard SMPL template, resulting in misalignment with visible parts and leading to inaccurate 3D reconstructions.
(\romannumeral2) Furthermore, creating a comprehensive and precise 3D representation from a single RGB image is an ill-posed problem as the inherently limited information.
As illustrated in~\cref{fig:fig_1}{\color{red}b}, some methods that use 3D representations rely on localized 3D joints as \textbf{local supervision},
ignoring the rich semantic relations between voxels across different scenes. 
These \enquote{local} contents (\ie, human joints) occupy only a small portion of the 3D space, while most voxels are often occupied by occlusions and background.
Consequently, the lack of explicit supervision for the entire 3D space makes it difficult to differentiate target humans from other semantically similar voxels, resulting in ambiguous 3D representations.

Therefore, to improve occluded human mesh recovery, we consider investigating a fusion framework that integrates 2D and 3D features for \textbf{2D$\&$3D alignment}, along with a \textbf{global supervision} strategy to obtain a semantically clear 3D feature space. 
By leveraging the complementary information from both 2D and 3D representations, the network could overcome the limitations of using only a single 2D representation, enabling obscured human parts to be detected in 3D representations and achieving 2D$\&$3D alignment. 
Given a global supervision strategy, we could explicitly supervise the entire 3D space to highlight the representation of target humans and distinguish them from other semantically similar voxels, resulting in a semantically clear 3D feature space.

Based on the above motivation, this paper proposes  a novel framework, 3D JOint contrastive learning with TRansformers (JOTR), for recovering occluded human mesh using a fusion of multiple representations as shown in~\cref{fig:fig_1}{\color{red}a}.
Unlike existing methods such as 3DCrowdNet~\cite{3dCrwodNet} and BEV~\cite{BEV} that employ 3D-aware 2D sampling techniques, JOTR integrates 2D and 3D features through transformers ~\cite{transformer} with attention mechanisms.
Specifically, \methodname utilizes an encoder-decoder transformer architecture to combine 3D local features (\ie, sampled 3D joint features) and 2D global features (\ie, flatten 2D features), enhancing both 2D and 3D alignment.
Besides, to obtain semantically clear 3D representations, the main objective is to strengthen and highlight the human representation while minimizing the impact of irrelevant features (\eg, occlusions and background). 
Accordingly, we propose a new approach, 3D joint contrastive learning (in~\cref{fig:fig_1}{\color{red}b}), that provides global and explicit supervision for 3D space to improve the similarity of semantically similar voxels (\ie, human joints), while maintaining discrimination from other voxels (\eg, occlusions). 
By carefully designing 3D joint contrast for 3D representations, \methodname can mitigate the effects of occlusion and acquire semantically meaningful 3D representations, resulting in accurate localization of 3D human joints and acquisition of meaningful 3D joint features.

We conduct extensive experiments on both standard  3DPW benchmark~\cite{3dpw} and occlusion benchmarks such as 3DPW-PC~\cite{3dpw,ROMP}, 3DPW-OC~\cite{CVPE_2020_OOH,3dpw}, 3DPW-Crowd~\cite{3dpw,3dCrwodNet}, 3DOH~\cite{CVPE_2020_OOH} and CMU Panoptic~\cite{CMUpanoptic}, and \methodname achieves state-of-the-art performance on these datasets.
Especially, \methodname outperforms the prior state-of-the-art method 3DCrowdNet~\cite{3dCrwodNet} by \textbf{6.1}  (PA-MPJPE), \textbf{4.9} (PA-MPJPE), and \textbf{5.3} (MPJPE)  on 3DPW-PC, 3DPW-OC, and 3DPW  respectively.
Moreover, we carry out comprehensive ablation experiments to demonstrate the effectiveness of our framework and 3D joint contrastive learning strategy.
Our contributions are summarized as follows:
\begin{itemize}
	\vspace{-2mm}
	\item We propose \methodname, a novel method for recovering occluded human mesh using a fusion of 2D global and 3D local features, which overcomes limitations caused by person-person and person-object occlusions and achieves 2D$\&$3D aligned results. \methodname achieves state-of-the-art results on both standard and occluded datasets, including 3DPW, 3DPW-PC, 3DPW-OC, 3DOH, CMU Panoptic, and 3DPW-Crowd.
	\vspace{-1mm}
	\item We develop a 3D joint contrastive learning strategy that  supervises the 3D space explicitly and globally to obtain semantically clear 3D representations, minimizing the impact of occlusions and adapting to more challenging scenarios with the help of cross-image contrast. 
\end{itemize}
\section{Related Work}

Based on the incorporation of a human body model~\cite{SMPL,SMPL-X:2019}, Deep Neural Network-based 3D Human Mesh Recovery methods~\cite{i2L_meshNet,HMR,SPIN,3dCrwodNet,BEV,CVPE_2020_OOH,zhang2021pymaf,ROMP,PARE,jiang2020multiperson,metro_lin,GraphCMR,mesh_graphormer,choi2020pose2mesh,dwivedi2021learning,guan2021bilevel,kissos2020beyond,kocabas2020vibe,li2021hybrik,tripathi2020posenet3d,OCHMR,shen2023global} can be divided into two categories. The first, SMPL-based approaches~\cite{HMR,SPIN,PARE,zhang2021pymaf,ROMP,BEV,kocabas2020vibe,jiang2020multiperson,shen2023global}, maps input pixels to SMPL parameters~\cite{SMPL} such as pose and shape and reconstructs meshes by SMPL models, while the second, SMPL-free methods~\cite{metro_lin,GraphCMR,mesh_graphormer}, directly maps raw pixels to 3D mesh vertices  without the assistance of SMPL models. 
In this paper, we mainly consider the first method as the implementation approach.

\noindent\textbf{Human Mesh Recovery.} 
Usually, human mesh recovery methods estimate 3D human mesh of a single person within a person bounding box, which is scaled to the same size. This allows us to assume that the distance between each individual and the camera is roughly equivalent in the cropped image patch. 
Early works~\cite{HMR,SPIN} employ spatial averaging on CNN features for obtaining global features and utilize Multi-Layer Perceptrons (MLPs) to regress SMPL parameters. 
However, global pooling is not suitable for achieving pixel-aligned results, leading to subpar performance in real-world scenarios.
PARE~\cite{PARE} proposes using part segmentation masks to enhance pixel alignments.
Zhang~\etal~\cite{CVPE_2020_OOH} make use of occlusion segmentation masks to allow the model to attend to the visible human body parts, which also helps to reconstruct complete human mesh.
OCHMR~\cite{OCHMR} employs load-global center maps to make the model regress the mesh of the referred person.
While these methods make progress in occluded human mesh recovery by enhancing the ability to represent 2D information, they overlook the 3D structural information.
3DCrowdNet~\cite{3dCrwodNet} and BEV~\cite{BEV} introduce 3D representations to locate human joints in 3D space.
However, these approaches also have limitations since they extract 2D CNN features in the corresponding region of located 3D joints, thereby overlooking the full potential of 3D representations.
Therefore, we design a fusion framework to integrate 2D and 3D features for mutual complementation.

\noindent\textbf{Multi-Modality Transformers.} 
Following the success of vision transformers in processing image~\cite{dosovitskiy2020ViT,detr,cheng2021per,liu2021swin} or video~\cite{arnab2021vivit,yang2021AOT,yang2022decoupling}, multi-modality transformers~\cite{lei2021detecting, li2020hero, kamath2021mdetr, li2021referring, zhang2022eatformer, zhang2023rethinking, zhang2021analogous, li2023seg, samtrack} are capable of processing input data from multiple modalities, such as text, image, audio, or video, in a single model. 
The attention mechanism is a key component of transformers, which enables them to selectively focus on relevant parts of the input sequence when generating the output.
Returning to the present task, it is worth noting that although there is only one visual modality, two distinct representations (\ie, 2D and 3D representations) are available.
Thus, we propose using transformers with attention mechanisms to integrate multi-representation features, rather than relying on global average pooling or joint feature sampling in CNN features.

\noindent\textbf{Contrastive Learning.} Contrastive learning~\cite{chen2020simple,he2020momentum,grill2020bootstrap,chen2020big} is a type of unsupervised learning that aims to learn a similarity metric between data samples. The goal of contrastive learning is to bring similar examples closer together in feature space while pushing dissimilar examples farther apart.
With regards to the current objective, in 3D space, human body joints occupy a relatively small proportion, with the majority of voxels in the space being occupied by other objects (\eg, another person's body, background elements, and empty elements).
This poses a significant challenge in learning a similarity metric between data samples in 3D space.
To address this issue, we propose a novel joint-based contrastive learning strategy inspired by the recent success of pixel contrastive learning in semantic segmentation~\cite{zhong2021pixel, wang2021exploring, alonso2021semi, hu2021region}, which enables the network to learn a clear similarity metric in 3D space.

\begin{figure*}[t]
    \vspace{-5mm}
        \centering
        \includegraphics[width=0.95\linewidth]{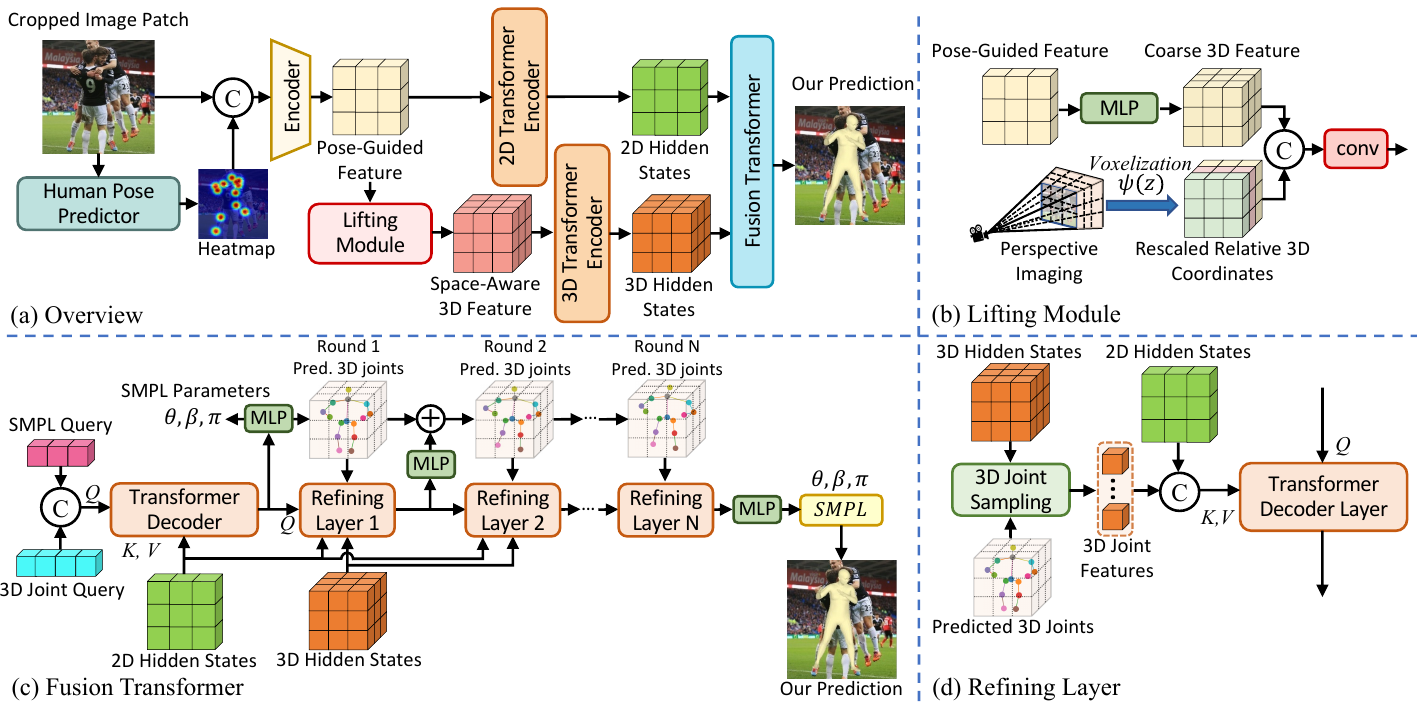}
        \vspace{-2.5mm}
        \caption{(a) The overview of our method. \methodname achieves 2D and 3D features from a cropped image patch and fuses them with a fusion transformer for 2D and 3D alignment. (b) The detail of the lifting module, which is responsible for lifting pose-guided 2D features to space-aware 3D features. (c) The fusion transformer is applied for fusing 2D and 3D features with attention mechanisms. (d) The refining layer combine sampled 3D joint features and 2D global features to refine the regression.
        }
        \label{fig:method_overview}
    \vspace{-5mm}
\end{figure*}

\section{Method}

\noindent\textbf{Human Body Model.} 
SMPL~\cite{SMPL} represents a 3D human mesh by 3 low-dimensional vectors (\ie, pose $\mathbf{\theta} \in \mathbb{R}^{72}$, shape $\mathbf{\beta} \in \mathbb{R}^{10}$ and camera parameters $\mathbf{\pi}  \in \mathbb{R}^{3}$).
Following previous methods~\cite{3dCrwodNet,ROMP,SPIN,HMR,PARE}, we use the gender-neutral shape model. 
The SMPL model generates a 3D mesh $\mathcal{M}(\theta,\beta) \in \mathbb{R}^{6890 \times 3}$ through a differentiable function.
By applying a pretrained linear regressor $W \in \mathbb{R}^{N\times6890}$,  we obtain the 3D joint coordinates $J_\mathit{3D}=W \mathcal{M} \in \mathbb{R}^{N\times3}$, where $N=17$, conveniently.
Additionally, we obtain the 2D joints $J_{2D} = \boldsymbol{\Pi}(J_{3D}, \pi) \in \mathbb{R}^{N \times 2}$  by projection.

\begin{figure}[t]
    \vspace{-3mm}
        \centering
        \includegraphics[width=0.99\linewidth]{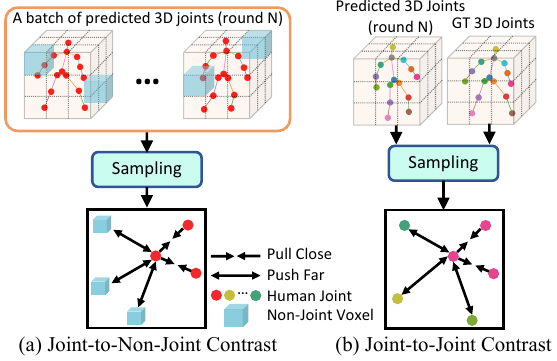}
        \vspace{-3mm}
        \caption{(a) The detail of joint-to-non-joint contrastive learning. (b) The detail of joint-to-joint contrastive learning.}
        \label{fig:contrast}
    \vspace{-4mm}
\end{figure}

\noindent\textbf{Overview.} 
We propose a method called \methodname, which utilizes transformers to fuse 2D and 3D features for 2D$\&$3D alignment and a novel contrastive learning strategy to globally supervise the 3D space for target humans.
Our pipeline is depicted in~\cref{fig:method_overview}{\color{red}a} and explained in~\cref{sec:fuse_2d_3d}, where JOTR regresses SMPL parameters by fusing 2D and 3D features obtained from a cropped image patch.
The proposed 3D joint contrastive learning is illustrated in~\cref{fig:contrast} and explained in~\cref{sec:joint_contrastive_learning}, including two contrastive losses: joint-to-non-joint contrast and joint-to-joint contrast.

\subsection{Fusing 2D and 3D Features with Transformers}
\label{sec:fuse_2d_3d}
As analyzed in~\cref{sec:introduction}, relying solely on 2D features for achieving 2D alignment to reconstruct the human mesh in occluded scenarios may result in suboptimal performance. To overcome this limitation, we propose integrating both 2D and 3D features with transformers in the reconstruction process. Drawing inspiration from the success of transformer models in multi-modality fusion~\cite{kamath2021mdetr,lei2021detecting}, we propose an  encoder-decoder transformer architecture that enables the mutual complementation of 2D and 3D features for 2D$\&$3D alignment.

\noindent\textbf{Lifting Module.} 
\label{sec:lifting_2d_to_3d}
Unlike previous method~\cite{3dCrwodNet} that lifts 2D features to 3D features via MLPs without integrating inductive bias or prior knowledge, we draw inspiration from bird's eye view representations~\cite{reiher2020sim2real, yang2021projecting, chitta2021neat, wang2022detr3d, li2022bevformer, BEV} in 3D space. 
As analyzed in BEV~\cite{BEV}, the farther a voxel is from the camera, the less information it carries. To put this hypothesis into practice, BEV employs pre-defined 3D camera anchor maps to impact the 3D feature.
Similar to BEV, we design learnable Rescaled Relative 3D Coordinates (RRC) $C_{3D} \in \mathbb{R}^{D \times H \times W \times 3}$ in range $(0, 1)$ to provide 3D spatial prior knowledge.
In this representation, $C_{ijk} \in \mathbb{R}^{3}$  represents the relative location of voxel
$(x_{k}, y_{j}, z_{i})$ and  the $x$ and $y$ coordinates are uniformly distributed with equal intervals. 
For $Z$ axis, we utilize a monotonically increasing convex function $\psi$ to rescale $z$ coordinates unevenly as $z' = \psi (z)$.
In practice, we employ $\psi(z) = z^{\lambda}, \lambda > 1$ as rescaling function and $\lambda$ is a learnable parameter with initial value of $3.0$. 
The whole pipeline can be written as:
\vspace{-2pt}
{
    \small
    \begin{align*}
        \small
        \label{eq:lifting_2d_to_3d}
        \hat{F_\mathit{3D}} &= MLP(F_{2D}) , \\
        \tilde{F_{3D}} &=   CNN(Concat(\hat{F_\mathit{3D}}, C_{3D})) , \\
        H_{3D} &= TransformerEncoder(\tilde{F_{3D}}).
    \vspace{-2pt}
    \end{align*}
}
\methodname first lifts pose-guided 2D feature $F_{2D} \in \mathbb{R}^{H \times W \times C}$ which is obtained from image and joint heatmap through CNN encoder to coarse 3D feature $\hat{F_\mathit{3D}} \in \mathbb{R}^{ D\times H\times W \times C}$ via MLPs without any inductive bias or prior knowledge. 
Then, \methodname concatenates $\hat{F_\mathit{3D}}$ and $C_{3D}$ in channel dimension.
Following CoordConv~\cite{liu2018coordconv}, we apply a convolutional block to refine the concatenated feature to achieve space-aware 3D feature $\tilde{F_\mathit{3D}} \in \mathbb{R}^{ D\times H\times W \times C }$. 
Finally, we utilize a transformer encoder (\ie, 3D transformer encoder in~\cref{fig:method_overview}{\color{red}a}) to enhance the global interaction of 3D space via self-attention mechanism, 
\vspace{-4pt}
\begin{equation}
    \small
    \label{eq:attention}
    Attention(Q, K, V)=softmax\left(\frac{Q K}{\sqrt{C}}\right) V,
\vspace{-2pt}
\end{equation}
achieving the hidden state $H_{3D} \in \mathbb{R}^{ D \times H \times W \times C }$.
For the sake of simplicity, we omit the positional encoding and rearrangement of tensor in~\cref{eq:attention}.

\noindent\textbf{Fusion Transformer.}
In contrast to prior 2D alignment technologies such as spatial averaging and 2D joint feature sampling, we propose the use of attention mechanisms to selectively focus on semantically distinct areas (\ie, visible human parts).
Moreover, to estimate hidden information for achieving 3D alignment, we extend 2D features with 3D joint feature sampling. 
Drawing inspiration from the successful fusion of image and text representations in MDETR~\cite{kamath2021mdetr} and Moment-DETR~\cite{lei2021detecting}, we design a transformer decoder-based fusion transformer to integrate 2D and 3D features and regress SMPL parameters in a coarse-to-fine manner leading to 2D $\&$ 3D alignment.

\noindent\textbf{SMPL/Joint Query.}
Instead of concatenating or pooling on 2D and 3D features, \methodname decouples the SMPL parameters and 2D/3D joint features  into separate query tokens, $Query \in \mathbb{R}^{N_q \times C}$,  comprising two distinct parts.
The $N_s$ tokens belong to SMPL token, where $N_s = 3$, and are responsible for regressing pose, shape and camera parameters $ \{ \theta, \beta, \pi \}$ respectively.
The remaining  $N_j = N_q - N_s$ tokens are responsible for locating 3D joints of the human and extracting corresponding 3D joint features, which refine the SMPL parameters and provide auxiliary supervision for the 3D space.

\noindent\textbf{2D-Based Initial Regression.}
As shown in~\cref{fig:method_overview}{\color{red}c}, we initially have no prior knowledge about the 3D joint locations. We regress the SMPL parameters and initial 3D joints with a transformer decoder reasoning in 2D hidden state $H_{2D} \in \mathbb{R}^{H \times W \times C} $ which is obtained by a transformer encoder (\ie, 2D transformer encoder in~\cref{fig:method_overview}{\color{red}a}) working on $F_{2D}$. 
$H_{2D}$ is set as $K$ and $V$, and $Query$ tokens are treated as $Q$ in~\cref{eq:attention}. 
Subsequently, we obtain initial predictions for pose, shape, camera parameters, and 3D joint coordinates via MLPs working on the output of transformer decoder.  

\noindent\textbf{Refining with 3D Features.}
To conserve computing resources, we avoid directly concatenating the hidden states (\ie, $H_{2D}$ and $H_{3D}$).  
Instead, we use the initial prediction of 3D joints $J_{3D}' \in \mathbb{R}^{N_j \times 3}$ as reference points to sample \enquote{local} 3D joint features $H_{J_{3D}}=\mathcal{F}\left(H_{3D}, J_{3D}'\right) \in \mathbb{R}^{N_j \times C}$ like~\cite{ddetr} in~\cref{fig:method_overview}{\color{red}d}, where $\mathcal{F}(\cdot)$ denotes feature sampling and trilinear interpolation. 
We then concatenate $H_{2D}$ with $H_{J_{3D}}$ and feed them into another transformer decoder (\ie, a stack of refining layers in~\cref{fig:method_overview}{\color{red}c}) as $K$ and $V$ in~\cref{eq:attention}.
Note that $Z$ axis is not uniform in our 3D space.
When sampling \enquote{local} 3D joint features, we also apply $\psi$ to rescale $z$ in $J_{3D}'$ as mentioned earlier. 
Since the refining process consists of several identical transformer decoder layers, we naturally consider utilizing the outputs of each layer $H_{d} \in \mathbb{R}^{L \times N \times C}$  as a cascade refinement,
\vspace{-2mm}
\begin{equation}
    \small
    \label{cascade_refine}
        \beta^{l+1}=\beta^{l}+MLP\left(\beta^{l}, H_{d}^{l}\right),
\vspace{-2mm}
\end{equation}
where $l$ denotes the $l$-th refining layer and $MLP\left(\beta^{l}, H_{d}^{l}\right)$ is responsible for learning the residual for correcting parameters via MLPs. Besides, we also regress and $J_{3D}'$ and the input of Vposer~\cite{SMPL-X:2019} with cascaded refinement as  shown above. 

\subsection{3D Joint Contrastive Learning} 
\label{sec:joint_contrastive_learning}
As analyzed in~\cref{sec:introduction}, due to the lack of explicit \enquote{global} supervision for 3D representations, the \enquote{local} 3D joint coordinates may not provide accurate enough supervision for the 3D features.
Especially  when the target person is obstructed by other individuals, similarities in their semantic appearances could result in confusion.
To address this challenge, we propose a 3D joint contrastive learning strategy inspired by the success of pixel contrastive learning in semantic segmentation~\cite{wang2021exploring}.
This approach enhances the representation of the target person while distinguishing them from other objects (\eg, other people, occlusions, and background).

\noindent\textbf{Vanilla Contrastive Learning.} 
In computer vision, contrastive learning was originally applied for unsupervised representation learning, where the goal is to minimize the distance between similar images (\ie, an image with its augmented version) while maximizing the distance between dissimilar images (\ie, an image with another image in training set) in an embedding space. 
Usually, InfoNCE~\cite{gutmann2010noise,oord2018representation} is used as the loss function for contrastive learning,
\vspace{-4pt}
\begin{equation}\small\label{eq:NCE}
\!\!\!\!\mathcal{L}^{\text{NCE}}_I\!=\!-\log\frac{\exp(\bm{i}\!\cdot\!\bm{i}^+/\tau)}{\exp(\bm{i}\!\cdot\!\bm{i}^+/\tau)
\!+\!\sum_{\bm{i}^-\in \mathcal{N}_I}\exp(\bm{i}\!\cdot\!\bm{i}^-/\tau)},\!\!
\vspace{-3pt}
\end{equation}
where $I$ is the anchor image, $\bm{i} \in \mathbb{R}^{C}$ is the representation embedding of $I$,  $\bm{i}^+$ is an embedding of a positive for $I$, $\mathcal{N}_I$ contains embeddings of negatives, `$\cdot$' denotes the inner (dot) product, and $\tau\!>\!0$ is a temperature hyper-parameter. Note that all the embeddings in the loss function are $\ell_2$-normalized.

\noindent\textbf{Joint-to-Non-Joint Contrast.} 
As shown in \cref{fig:contrast}{\color{red}a}, to better distinguish occlusion cases, we consider \textit{joint-to-non-joint contrast} between the $n$-th round predicted joints (in~\cref{fig:method_overview}{\color{red}c}) and the entire 3D space, as there are many voxels outside the joints.
We augment~\cref{eq:NCE} in our joint-to-non-joint contrast setting.
Since we employ trilinear interpolation to acquire the joint embedding from $H_{3D}$, the joint embedding is a weighted sum of the 8 voxel embeddings in the 3D space.
As a result, for an anchor joint $j$, the positive samples are other predicted joints (not restricted to belonging to the same class), and the negative samples are the voxels that have no contribution to any joint embeddings. 
The joint-to-non-joint contrastive loss is defined as:
\begin{equation}
    \small
    \label{eq:j2v_NCE}
    \!\!\!\!\mathcal{L}^{\text{NCE}}_{j2n}\!=\!\frac{1}{|\mathcal{P}_j|}\!\!\sum_{\bm{j}^+\in\mathcal{P}_j\!\!}\!\!\!-_{\!}\log\frac{\exp(\bm{j}\!\cdot\!\bm{j}^{+\!\!}/\tau)}{\exp(\bm{j}\!\cdot\!\bm{j}^{+\!\!}/\tau)
    +\!\sum\nolimits_{\bm{n}^{-\!}\in\mathcal{N}_n\!}\!\exp(\bm{j}\!\cdot\!\bm{n}^{-\!\!}/\tau)},\!\!
\end{equation}
where $\mathcal{P}_j$ is  joint embedding collections of positive samples and $\mathcal{N}_n$ denote non-joint voxel embedding collections of negative samples, for joint $j$.

\noindent\textbf{Joint-to-Joint Contrast.} 
As shown in \cref{fig:contrast}{\color{red}b}, to strengthen the internal connections among joints of the same category, we consider \textit{joint-to-joint contrast} among human joints.
We extend~\cref{eq:NCE} for applying to our joint-to-joint contrast setting.
Essentially, the data samples in our contrastive loss computation are the  $n$-th round predicted joints  (in~\cref{fig:method_overview}{\color{red}c}) and ground truth 3D joints.
For an anchor joint $j$ from predicted joints with its corresponding semantic label $\bar{c}$ (\eg, head, right hand, and neck), the positive samples are ground truth joints that also belong to the class $\bar{c}$, and the negative samples are the $n$-th round predicted joints belonging to the other classes $\mathcal{C}\setminus\{c_j\}$.
As a result, the joint-to-joint contrastive loss is defined as:
\begin{equation}
    \small
    \label{eq:j2j_NCE}
    \!\!\!\!\mathcal{L}^{\text{NCE}}_{j2j}\!=\!\frac{1}{|\mathcal{P}_j|}\!\!\sum_{\bm{j}^+\in\mathcal{P}_j\!\!}\!\!\!-_{\!}\log\frac{\exp(\bm{j}\!\cdot\!\bm{j}^{+\!\!}/\tau)}{\exp(\bm{j}\!\cdot\!\bm{j}^{+\!\!}/\tau)
    +\!\sum\nolimits_{\bm{j}^{-\!}\in\mathcal{N}_j\!}\!\exp(\bm{j}\!\cdot\!\bm{j}^{-\!\!}/\tau)},\!\!
\end{equation}
where $\mathcal{P}_j$ and $\mathcal{N}_j$ denote joint embedding collections of the positive and negative samples, respectively, for joint $j$. 

Note that the positive and negative samples, as well as the anchor joint $j$ in both \textit{joint-to-non-joint} and \textit{joint-to-joint} contrast are not necessarily limited to the same 3D space. 
The joint-to-non-joint contrastive loss in~\cref{eq:j2v_NCE} and joint-to-joint contrastive loss in~\cref{eq:j2j_NCE} are complementary to each other; the former enables the network to learn discriminative joint features that are distinctly different from those of other non-joint voxels (\eg, occlusions), 
while the latter helps to regularize the joint embedding space by improving intra-class compactness and inter-class separability.

\subsection{Loss Function.}
Finally, we obtain refined SMPL parameters $\{ \theta, \beta, \pi \}$. 
We can achieve mesh vertices $M=\mathcal{M}(\theta,\beta) \in \mathbb{R}^{6890 \times 3}$ and 3D joints from mesh $J_\mathit{3D}=W \mathcal{M} \in \mathbb{R}^{N\times3}$ accordingly. 
We follow common practices~\cite{HMR,SPIN,3dCrwodNet} to project 3D joints on 2D space  $J_{2D} = \boldsymbol{\Pi}(J_{3D}, \pi) \in \mathbb{R}^{N \times 2}$ and  add supervisions with 2D keypoints.
Meanwhile, when 3D annotations are available, we also add 3D supervision on SMPL parameters and 3D joint coordinates. 
Overall, the loss function can be written as follows:
\vspace{-2pt}
\begin{equation}
    \small
    \label{eq:loss}
    \begin{aligned}
        \mathcal{L} = &\lambda_{3 D} \mathcal{L}_{3 D}+\lambda_{2 D} \mathcal{L}_{2 D}+\lambda_{SMPL} \mathcal{L}_{S M P L} \\
                    &+ \lambda_{j2n} \sum\nolimits_j \mathcal{L}^{\text{NCE}}_{j2n} + \lambda_{j2j} \sum\nolimits_j \mathcal{L}^{\text{NCE}}_{j2j},
    \end{aligned}
\end{equation}
where $j$ is the sampled anchor joints and the first three is calculated as:
\vspace{-2pt}
{
    \small
    \begin{align*}
        \mathcal{L}_{\mathit{3D}} & =  \| J_{3D} \; - \; \hat{J_{3D}}\| , \\
        \mathcal{L}_{\mathit{2D}} &=  \| J_{2D} \; - \; \hat{J_{2D}} \| , \\
        \mathcal{L}_{\mathit{SMPL}} &= \| \theta \; - \; \hat{\theta} \| + \|\beta \; - \; \hat{\beta}\|,
    \vspace{-4pt}
    \end{align*}
}
where $\| \cdot \|$ denotes L1 norm. $\hat{J_{2D}}$, $\hat{J_{3D}}$, $\hat{\theta}$, and $\hat{\beta}$ denote the
ground truth 2D keypoints, 3D joints, pose parameters and shape parameters, respectively. 
\begin{table*}[t]
  \vspace{-2mm}
    \centering
    \footnotesize
    \setlength{\tabcolsep}{1pt} %
    \renewcommand{\arraystretch}{1.15} 
      \begin{tabularx}{\linewidth}{>{\raggedright\arraybackslash}p{2.1cm}|>{\centering\arraybackslash}X>{\centering\arraybackslash}p{1.55cm}>{\centering\arraybackslash}X | >{\centering\arraybackslash}X>{\centering\arraybackslash}p{1.55cm} >{\centering\arraybackslash}X | >{\centering\arraybackslash}X>{\centering\arraybackslash}p{1.55cm}>{\centering\arraybackslash}X | >{\centering\arraybackslash}X>{\centering\arraybackslash}p{1.55cm}>{\centering\arraybackslash}X}
    \Xhline{3\arrayrulewidth}
   
    \multirow{2}{*}{\textbf{Method}} & \multicolumn{3}{c|}{\textbf{3DPW-OC} } & \multicolumn{3}{c|}{\textbf{3DOH} } & \multicolumn{3}{c|}{\textbf{3DPW-PC }} & \multicolumn{3}{c}{\textbf{3DPW-Crowd }}   \\
    
      & \textbf{MPJPE$\downarrow$} & \textbf{PA-MPJPE$\downarrow$} & \textbf{PVE$\downarrow$} & \textbf{MPJPE$\downarrow$} & \textbf{PA-MPJPE$\downarrow$} & \textbf{PVE$\downarrow$} & \textbf{MPJPE$\downarrow$} & \textbf{PA-MPJPE$\downarrow$} & \textbf{PVE$\downarrow$}& \textbf{MPJPE$\downarrow$} & \textbf{PA-MPJPE$\downarrow$} & \textbf{PVE$\downarrow$}\\
    \hline
    I2L-MeshNet~\cite{i2L_meshNet}  & 92.0 & 61.4 & 129.5 & - & - & - & 117.3 & 80.0 & 160.2 & 115.7 & 73.5 & 162.0   \\
    SPIN~\cite{SPIN} & 95.5 & 60.7 & 121.4 & 110.5 & 71.6 & 124.2 & 122.1 & 77.5 & 159.8  & 121.2 & 69.9 & 144.1   \\
    PyMAF~\cite{zhang2021pymaf}  & 89.6 & 59.1 & 113.7 & 101.6 & 67.7 & 116.6 & 117.5 & 74.5 & 154.6 & 115.7 & 66.4 & 147.5   \\
    ROMP~\cite{ROMP}  & 91.0 & 62.0  & -  & - & - & - & 98.7 & 69.0 & - & 104.8 & 63.9 & 127.8   \\ 
    OCHMR~\cite{OCHMR}  & 112.2 & 75.2 &  145.9 & - & - & - & - & - & - & - & - & -  \\
    PARE*~\cite{PARE} & 83.5 & 57.0 & 101.5  & 109.0 & 63.8 & 117.4 &  96.8 & 64.5 & 122.4  &  94.9 & 57.5 & 117.6  \\
    3DCrowdNet~\cite{3dCrwodNet}  & 83.5 & 57.1 & 101.5 & 102.8 & 61.6 & 111.8 & 90.9 & 64.4 & 114.8 & 85.8 & 55.8 & 108.5  \\
    
    \hline
    Ours & \textbf{75.7} & \textbf{52.2}  & \textbf{92.6} & \textbf{98.7} & \textbf{59.3} & \textbf{104.8}& \textbf{86.5} &\textbf{58.3} & \textbf{109.7} & \textbf{82.4} & \textbf{52.0} &\textbf{103.4}\\
    \Xhline{3\arrayrulewidth}
    \end{tabularx}
    \vspace{-3mm}
    \caption{Comparisons to the state-of-the-art methods under severe occlusion. The units for mean joint and vertex errors are in mm.  PARE* use a HRNet-32 backbone, others are with ResNet-50. }
    \label{table:oc_sota}
    \vspace{-5mm}
\end{table*}

\begin{table*}[t]
\vspace{-3mm}
    \setlength\tabcolsep{1.5mm}
    \parbox{0.47\linewidth}{
    \centering
        \footnotesize
        \vspace{2.5mm}
        \renewcommand{\arraystretch}{1.15} 
        \begin{tabularx}{\linewidth}{>{\raggedright\arraybackslash}p{2.4cm} |>{\centering\arraybackslash}X >{\centering\arraybackslash}p{1.6cm} >{\centering\arraybackslash}X}
            \Xhline{3\arrayrulewidth}
            \textbf{Method}  &\textbf{MPJPE$\downarrow$} & \textbf{PA-MPJPE$\downarrow$} &  \textbf{PVE$\downarrow$}\\
            \hline 
            HMR~\cite{HMR}  & 130.0 & 76.7 & - \\
            Kanazawa \etal~\cite{kanazawa2019learning}  & 116.5 & 72.6 & 139.3 \\
            GCMR~\cite{GraphCMR}  & - & 70.2 & - \\
            DSD-SATN~\cite{sun2019human}  & - & 69.5 & -\\
            SPIN~\cite{SPIN}  & 96.9 & 59.2 & 116.4\\
            
            I2L-MeshNet~\cite{i2L_meshNet}  & 93.2 & 58.6 & 136.5\\
            PyMAF~\cite{zhang2021pymaf}  & 92.8 & 58.9 & 110.1 \\
            OCHMR~\cite{OCHMR} & 89.7 & 58.3 & 107.1 \\
            EFT~\cite{joo2021exemplar_EFT}  & - & 54.2 & -\\
            ROMP~\cite{ROMP}  & 89.3 & 53.5 & 105.6\\
            PARE~\cite{PARE} & 82.9 & 52.3 & 99.7 \\
            3DCrowdNet~\cite{3dCrwodNet} & 81.7 & 51.2 & 98.3 \\
            \hline
            Ours  & \textbf{76.4}  & \textbf{48.7} & \textbf{92.6}\\
                \Xhline{3\arrayrulewidth}
    \end{tabularx}
        \vspace{-2mm}
        \caption{Comparisons to the state-of-the-art methods on standard 3DPW \cite{3dpw} test split.}
        \label{table:3dpw}}
    \hspace{8.7mm}
    \parbox{.47\linewidth}{
    \centering
    \footnotesize
    \vspace{1mm}
    \renewcommand{\arraystretch}{1.15} 
    \begin{tabularx}{\linewidth}{>{\raggedright\arraybackslash}p{2.2cm}| >{\centering\arraybackslash}X >{\centering\arraybackslash}X >{\centering\arraybackslash}X >{\centering\arraybackslash}X | >{\centering\arraybackslash}X }
        \Xhline{3\arrayrulewidth}
        \textbf{Method}  &\textbf{Haggl.} & \textbf{Mafia} &  \textbf{Ultim.} & \textbf{Pizza} & \textbf{Mean} \\
        \hline 
        Zanfir~\etal~\cite{zanfir2018monocular} & 140.0 & 156.9 & 150.7 & 156.0 & 153.4 \\
        Zanfir\etal~\cite{zanfir2018deep} & 141.4 & 152.3 & 145.0 & 162.5 & 150.3 \\
        Jiang~\etal~\cite{jiang2020multiperson} & 129.6 & 133.5 & 153.0 & 156.7 & 143.2 \\
        ROMP~\cite{ROMP} & 111.8 & 129.0 & 148.5 & 149.1 & 134.6 \\
        SPIN~\cite{SPIN} & 124.3 & 132.4 & 150.4 & 153.5 & 133.1 \\
        OCHMR~\cite{OCHMR} & 115.5 & 123.7 & 142.6 & 150.6 & 133.1 \\
        REMIPS~\cite{fieraru2021remips} & 121.6 & 137.1 & 146.4 & 148.0 & 138.3 \\
        3DCrowdNet~\cite{3dCrwodNet} & 109.6 & 135.9 & 129.8 & 135.6 & 127.6 \\
        BEV*~\cite{BEV} & 110.3 & 125.6 & 150.7 & 131.7 & 127.9 \\
        \hline
        Ours & \textbf{99.9} & \textbf{113.5} & \textbf{115.7} & \textbf{123.6} & \textbf{114.7}\\
            \Xhline{3\arrayrulewidth}
    \end{tabularx}
    \vspace{-1mm}
    \caption{Comparison on CMU-Panoptic \cite{CMUpanoptic}. The numbers denote MPJPE. For a fair comparison, we apply BEV* model that is not fine-tuned on AGORA \cite{patel2021agora} (\ie, a synthetic 3D dataset.).}  \label{table:cmu}}
\vspace{-4mm}
\end{table*}\vspace{-1mm} 

\begin{table*}[t]
    \setlength\tabcolsep{1.5mm}
    \parbox{0.47\linewidth}{
    \centering
        \footnotesize
        \vspace{0mm}
        \renewcommand{\arraystretch}{1.15} 
            \begin{tabularx}{\linewidth}{>{\centering\arraybackslash}X |>{\centering\arraybackslash}X |>{\centering\arraybackslash}X  >{\centering\arraybackslash}p{1.65cm}  >{\centering\arraybackslash}X  }
                \Xhline{3\arrayrulewidth}
                2D Feature & 3D Feature & \textbf{MPJPE $\downarrow$} & \textbf{PA-MPJPE$\downarrow$} & \textbf{PVE$\downarrow$} \\
                \hline
                sampling & none  & 78.3 & 54.2 & 94.7  \\ 
                flatting & none & 77.8 & \textbf{53.2} & 94.7  \\ 
                sampling & sampling & 77.4 & 53.8 & 94.6  \\   
                \hline
                flatting & sampling & \textbf{77.0} & 53.6 & \textbf{94.3}  \\ 
                \Xhline{3\arrayrulewidth}
             \end{tabularx}
             \vspace{-3mm}
             \caption{Ablation study of utilization of 2D and 3D features. Flatting: flatting in height and weight for tokenization. Sampling: joint feature sampling for tokenization.}\label{tab:2d_3d}
    }
    \hspace{8.7mm}
    \parbox{.47\linewidth}{
    \centering
    \footnotesize
    \renewcommand{\arraystretch}{1.15} 
    \begin{tabularx}{\linewidth}{>{\centering\arraybackslash}p{2.9cm}|>{\centering\arraybackslash}X  >{\centering\arraybackslash}p{1.7cm} >{\centering\arraybackslash}X }
           \Xhline{3\arrayrulewidth}
           Index of Refining Layer & \textbf{MPJPE $\downarrow$} & \textbf{PA-MPJPE$\downarrow$} & \textbf{PVE$\downarrow$} \\
           \hline
           0 & 276.5 & 124.7 & 308.8  \\ 
           1   & 145.5 & 103.2 & 185.9  \\ 
           2  & 109.8 & 68.7 & 131.6  \\ 
           3 & 77.0 & 53.6 & 94.3  \\ 
           \Xhline{3\arrayrulewidth}
        \end{tabularx}
        \vspace{-3mm}
        \caption{Validation of coarse-to-fine regression. We take intermediate outputs of refining layers for regressing SMPL parameters. Zero stands for 2D-based initial regression.}\label{tab:coarse_to_fine}
    }

    \parbox{.47\linewidth}{
    \centering
    \footnotesize
    \vspace{3mm}
    \renewcommand{\arraystretch}{1.15} 
    \begin{tabularx}{\linewidth}{>{\centering\arraybackslash}p{2.5cm} |>{\centering\arraybackslash}X  >{\centering\arraybackslash}p{1.7cm}  >{\centering\arraybackslash}X  }
           \Xhline{3\arrayrulewidth}
           w/o SMPL Token & \textbf{MPJPE $\downarrow$} & \textbf{PA-MPJPE$\downarrow$} & \textbf{PVE$\downarrow$} \\
           \hline
           \xmark & 79.8 & 54.0 & 95.9 \\ 
           \cmark & \textbf{77.0} & \textbf{53.6} & \textbf{94.3}   \\ 
           \Xhline{3\arrayrulewidth}
        \end{tabularx}
        \caption{Ablation study of decoupling SMPL query. We apply average pooling on outputs of joint query tokens for regressing SMPL parameters when without SMPL query.}\label{tab:smpl_token}
    
    }
    \hspace{8.7mm}
    \parbox{.47\linewidth}{
    \centering
    \footnotesize
    \vspace{1mm}
    \renewcommand{\arraystretch}{1.2} 
    \begin{tabularx}{\linewidth}{>{\centering\arraybackslash}X|>{\centering\arraybackslash}X|>{\centering\arraybackslash}X>{\centering\arraybackslash}p{1.7cm} >{\centering\arraybackslash}X }
        \Xhline{3\arrayrulewidth}
        J2N & J2J & \textbf{MPJPE $\downarrow$} & \textbf{PA-MPJPE$\downarrow$} & \textbf{PVE$\downarrow$} \\
        \hline
        \xmark & \xmark & 77.0 & 53.6 & 94.3  \\ 
        \cmark & \xmark & 75.9 & 52.5 & 93.0  \\ 
        \xmark & \cmark & 76.6 & 52.2 & 93.2 \\ 
        \hline
        \cmark & \cmark & \textbf{75.7} & \textbf{52.2} & \textbf{92.6}  \\ 
        \Xhline{3\arrayrulewidth}
     \end{tabularx}
     \vspace{-2mm}
     \caption{Ablation study of 3D joint contrastive learning. J2N: joint-to-non-joint contrast. J2J: joint-to-joint contrast.}\label{tab:contrastive}
    }
\vspace{-4mm} 
\end{table*}

\section{Experiments}
\label{sec:experiments}

\noindent\textbf{Implementation Detail.}
This proposed \methodname is validated on the ResNet-50~\cite{resnet} backbone. Following 3DCrowdNet~\cite{3dCrwodNet}, we initialize ResNet from Xiao~\etal~\cite{xiao2018simple} for fast convergence. We use AdamW optimizer~\cite{AdamW} with a batch size of 256 and weight decay of $10^{-4}$. The initial learning rate is  $10^{-4}$. 
The ResNet-50 backbone takes a $256 \times 256$ image as input and produces image features with size of $2048 \times 8 \times 8 $. We build the 3D features with size of $256 \times 8 \times 8 \times 8$ and 2D features with size of $256 \times 8 \times 8$. 
As for weights for multiple different losses, we follow~\cite{Kendall_2018_CVPR} to adjust them dynamically using learnable parameters.
For joint-to-non-joint contrast, we sample $100$ anchor joints per GPU in each mini-batch, which are paired with $1024$ positive and $2048$  and negative samples. For joint-to-joint contrast, we sample $100$ anchor joints per GPU in each mini-batch, which are paired with $128$ positive and $256$ and negative samples. Both contrastive losses are set to a temperature of $0.07$.
More details can be found in the supplementary material. 

\noindent\textbf{Training.} Following the settings of previous work~\cite{HMR,SPIN,3dCrwodNet}, our approach is trained on a mixture of data from several datasets with 3D and 2D annotations, including Human3.6M~\cite{ionescu_h36m}, MuCo-3DHP~\cite{muco}, MSCOCO~\cite{coco},  and CrowdPose~\cite{li2018crowdpose}. Only the training sets are used, following the standard split protocols. For the 2D datasets, we also utilize their pseudo ground-truth SMPL parameters~\cite{Moon_2022_CVPRW_NeuralAnnot} for training. 

\noindent\textbf{Evaluation.} The 3DPW~\cite{3dpw} test split, 3DOH~\cite{CVPE_2020_OOH} test split, 3DPW-PC~\cite{ROMP,3dpw}, 3DPW-OC~\cite{CVPE_2020_OOH,3dpw}, 3DPW-Crowd~\cite{3dCrwodNet,3dpw} and CMU-Panoptic~\cite{CMUpanoptic} datasets are used for evaluation.
3DPW-PC and 3DPW-Crowd are the \textit{person-person} occlusion subset of 3DPW, 3DPW-OC is the \textit{person-object} occlusion subset of 3DPW and 3DOH is another \textit{person-object} occlusion specific dataset.
We adopt per-vertex error (PVE) in mm to evaluate the 3D mesh error. 
We employ Procrustes-aligned mean per joint position error (PA-MPJPE) in mm and mean per joint position error (MPJPE) in mm to evaluate the 3D pose accuracy. As for CMU-Panoptic, we only report  mean per joint position error (MPJPE) in mm following previous work~\cite{jiang2020multiperson, 3dCrwodNet, ROMP}.

\subsection{Comparison to the State-of-the-Art on Occlusion Benchmark}
\noindent\textbf{3DPW-OC~\cite{3dpw,CVPE_2020_OOH}} is a person-object occlusion subset of 3DPW and contains 20243 persons.~\cref{table:oc_sota} shows our method  achieve a new state-of-the-art performance on 3DPW-OC.

\noindent\textbf{3DOH~\cite{CVPE_2020_OOH}} is a person-object occlusion-specific dataset and contains 1290 persons in testing set, which incorporates a greater extent of occlusions than 3DPW-OC. 
For a fair comparison, we initialize PARE with weights that are not trained on the 3DOH training set, resulting in different performances from the results reported in~\cite{PARE}.
\cref{table:oc_sota} shows our method surpasses all the competitors with $59.3$  (PA-MPJPE). 

\noindent\textbf{3DPW-PC~\cite{3dpw,ROMP}} is a multi-person subset of 3DPW and contains 2218 persons' annotations under person-person occlusion.~\cref{table:oc_sota} shows our method surpasses all the competitors with $58.8$  (PA-MPJPE).

\noindent\textbf{3DPW-Crowd~\cite{3dpw, 3dCrwodNet}} is a person crowded  subset of 3DPW and contains 1923 persons.  We slightly surpass previous state-of-the-art as shown in~\cref{table:oc_sota}.
 
\noindent\textbf{CMU-Panoptic~\cite{CMUpanoptic}} is a dataset with multi-person indoor scenes. We follow previous methods~\cite{3dCrwodNet,jiang2020multiperson} applying 4 scenes for evaluation without using any data from training set. 
\cref{table:cmu}, shows that our method outperforms previous 3D human pose estimation methods on CMU-Panoptic, which means our model also works well for indoor and daily life scenes.
\vspace{-2mm}
\subsection{Comparison to the State-of-the-Art on Standard Benchmark}
\vspace{-2mm}
\noindent\textbf{3DPW~\cite{3dpw}} is the latest large-scale benchmark for 3D human mesh recovery. We do not use the training set and report performance on its test split which contains 60 videos and 3D annotations of 35515 persons. As shown in~\cref{table:3dpw}, Our method achieves  state-of-the-art  results among previous approaches.
 The results demonstrate the robustness of \methodname to a variety of in-the-wild scenarios.
\vspace{-2mm}
\subsection{Analysis.}
\vspace{-2mm}
In this section, we analyze the main components of \methodname and evaluate their impact on the mesh recovery performance.
More details and ablation studies can be found in the supplementary material.

\begin{figure}[t]
	\centering
	\vspace{-6mm}
	\includegraphics[width=\linewidth]{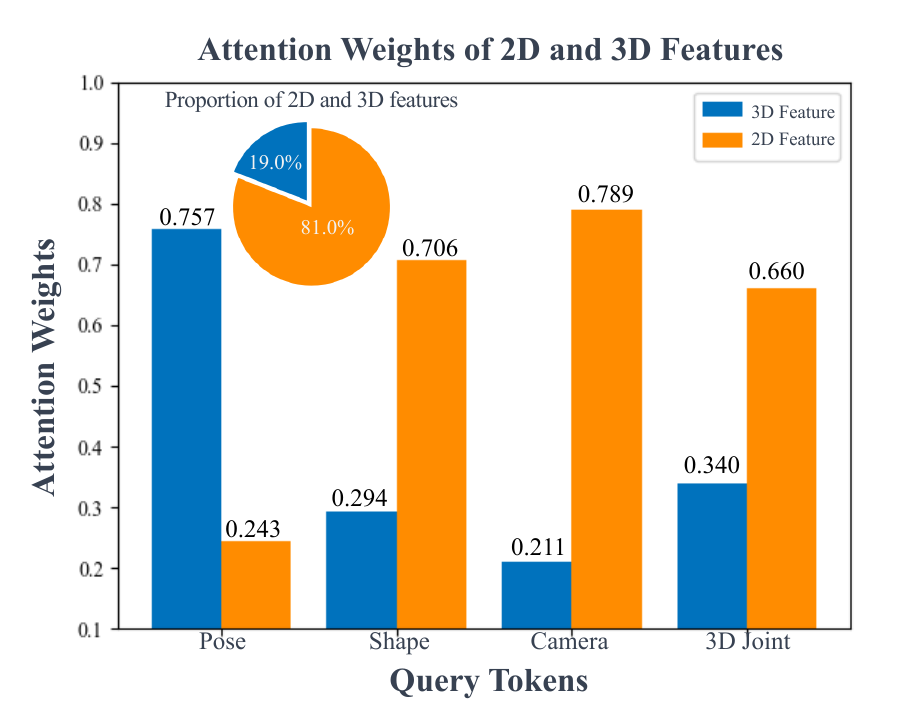}
	\vspace{-8mm}
    \caption{Visualization of cross-attention weights in the last refining layer. We randomly sample 100 persons from 3DPW test set and average the attention weights for visualization.}\label{fig:_weight}
	\vspace{-3mm}
\end{figure}

\noindent\textbf{Utilization of 2D and 3D features:}
\cref{tab:2d_3d} demonstrates that the incorporation of 3D features is beneficial for mesh recovery performance. For the utilization of 2D features, flatting shows better performance than sampling, which supports our hypothesis that sampling joint features in obscured regions could have a negative impact.
For 3D features, we do not conduct experiments for flatting 3D features due to memory limitations. Moreover, we believe that 3D joint feature sampling is adequate for alleviating occlusion problems by attending to the accurate depth.
\cref{fig:_weight} shows the attention weights in the last refining layer. The query tokens significantly pay more attention to 3D features, which validates the usefulness of our fusion framework.

\noindent\textbf{Validation of coarse-to-fine regression:}
We validate the accuracy of intermediate predictions of fusion transformer in~\cref{tab:coarse_to_fine}, which shows the coarse-to-fine regression process in \methodname.

\noindent\textbf{Decoupling SMPL query:}
\methodname performance improvement is observed in~\cref{tab:smpl_token} by decoupling SMPL query from joint query. In the experiment without decoupling, we employ mean pooling on the decoder's output and regress SMPL parameters through MPLs. Decoupling SMPL query is presumed to enhance performance by reducing interference in executing other tasks (\eg, joint localization) during SMPL parameter regression.

\begin{figure}[t]
	\centering
	\vspace{-4mm}
	\includegraphics[width=\linewidth]{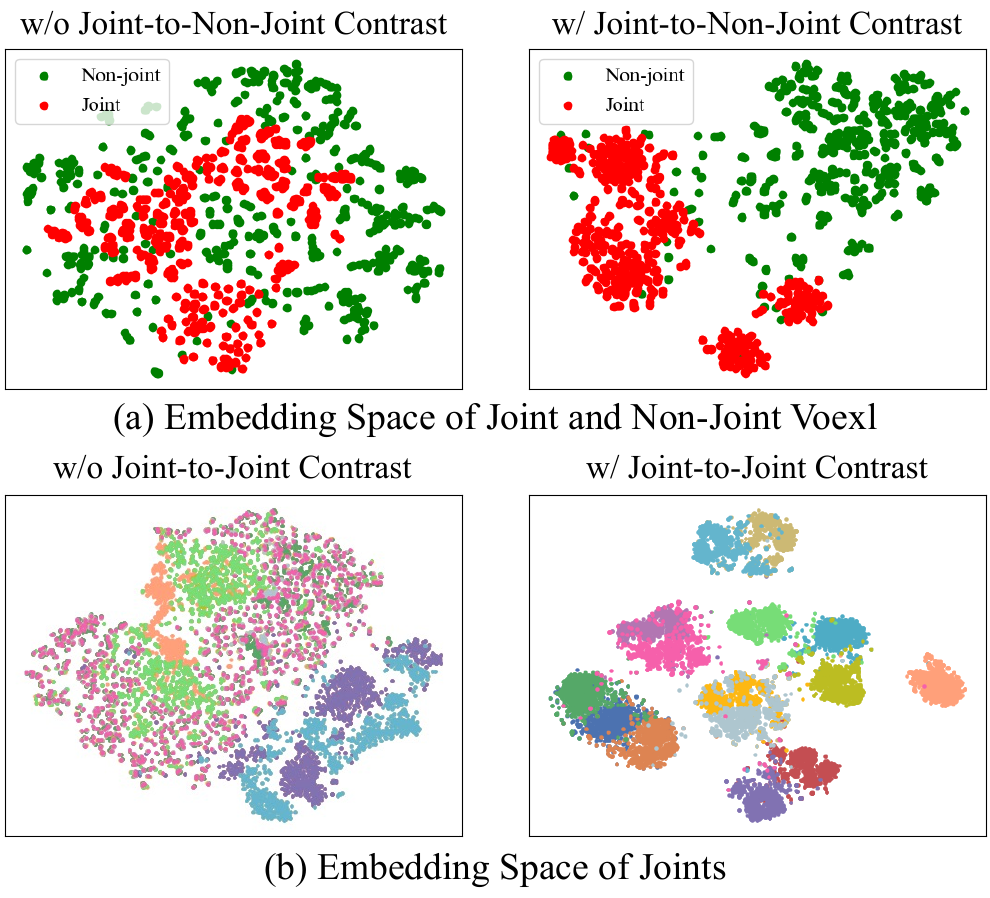}
	\vspace{-7mm}
    \caption{
		Visualization of features learned with (left) “local”
		joint supervision and (right) our “global” 3D joint contrast
		optimization objective (i.e., ~\cref{eq:j2v_NCE} and~\cref{eq:j2j_NCE}) on 3DPW
		test set~\cite{3dpw}. 
		Each color stands for a kind of joint (\eg, head and right knee) in (b).
		}
	\label{fig:joint_contrast}
	\vspace{-1mm} 
\end{figure}

\begin{figure}[t]
    \centering
    \vspace{-3mm}
	\includegraphics[width=0.95\linewidth]{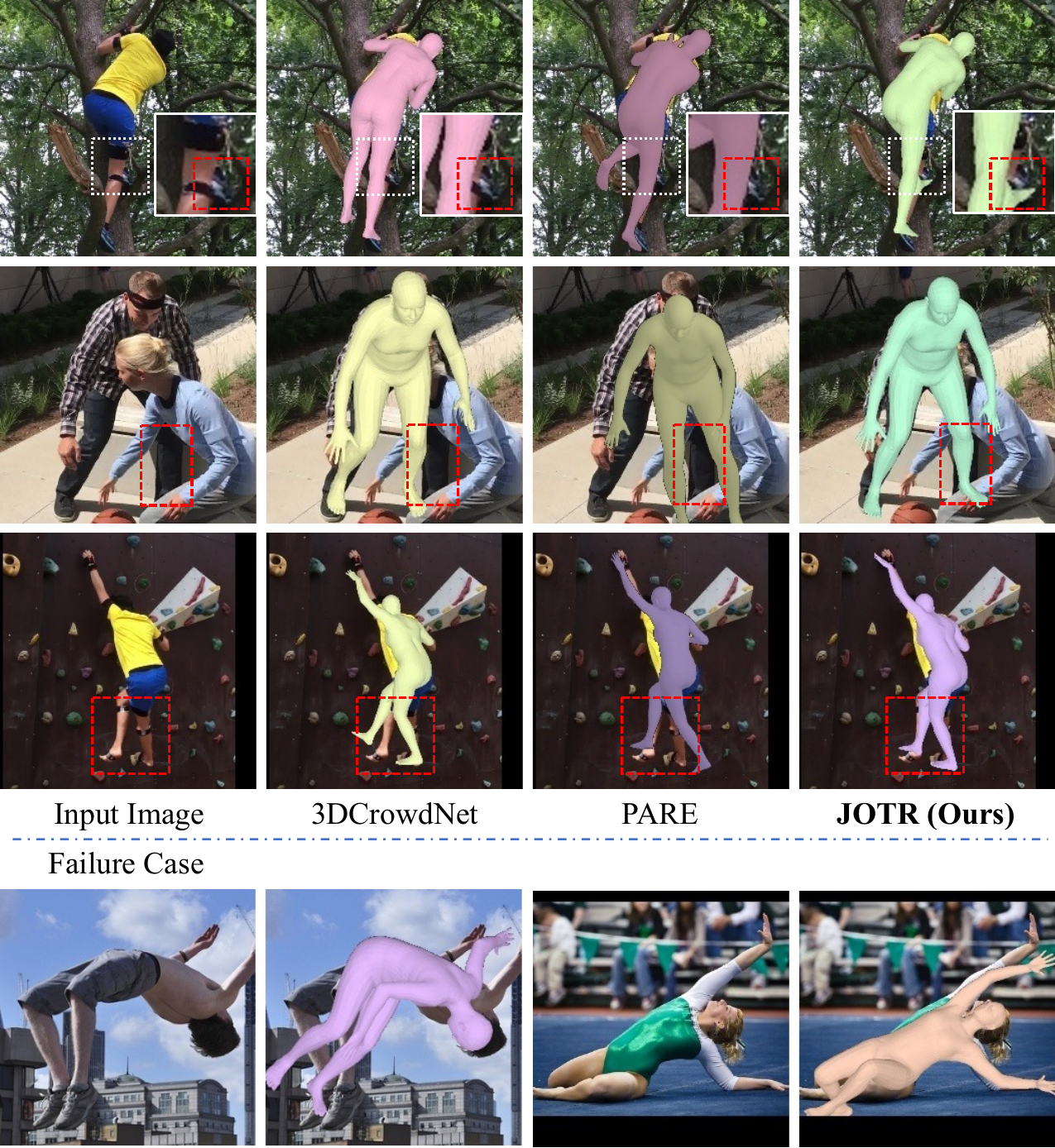}
    \vspace{-2mm}
    \caption{Qualitative results on 3DPW dataset~\cite{3dpw}. Note that we use no data from 3DPW for training. The bottom row shows
	the failure cases of \methodname. \methodname performs badly on extreme poses due to the lack of training data. More qualitative results can be found in the supplementary material.}
    \label{fig:comapre}
	\vspace{-5mm}
\end{figure}

\noindent\textbf{3D Joint contrastive learning:}
The impact of 3D joint contrastive learning on the performance of \methodname is presented in~\cref{tab:contrastive}. Both joint-to-non-joint and joint-to-joint contrastive losses result in improved performance, with the former being more effective as it incorporates global supervision for the entire 3D space. 
Our contrastive losses also lead to more compact and well-separated 
learned joint embeddings, as shown in~\cref{fig:joint_contrast}. 
This indicates that our network can generate more discriminative 3D features, producing semantically clear 3D spaces and promising results.

\vspace{-2mm}
\section{Conclusion}
\vspace{-2mm}
Many human mesh recovery methods focus on 2D alignment technologies, which would fail under occlusions or limited visibility. To address this limitation, we propose JOTR, a novel method that combines 2D and 3D features using an encoder-decoder transformer architecture to achieve 2D$\&$3D alignment.  Furthermore, we introduce two noevl 3D joint contrastive losses that enable global supervision of the 3D space of target persons, producing meaningful 3D representations.
Extensive experiments on 3DPW benchmarks show that JOTR achieves the new state of the art.

\noindent\textbf{Limitations and Broader Impact.} 1) JOTR relies on the human pose predictor to detect 2D keypoints, leading to long inference times. 2)  In the future, JOTR has the potential to be integrated with bottom-up 3D human mesh recovery methods for real-time applications.

\noindent\textbf{Acknowledgements.} This work was supported by the Natural Science Foundation of Zhejiang Province (DT23F020008) and the Fundamental Research Funds for the Central Universities~(226-2023-00051). 

{\small
\bibliographystyle{ieee_fullname}
\bibliography{main}

\begin{thebibliography}{10}\itemsep=-1pt

\bibitem{alonso2021semi}
Inigo Alonso, Alberto Sabater, David Ferstl, Luis Montesano, and Ana~C Murillo.
\newblock Semi-supervised semantic segmentation with pixel-level contrastive
  learning from a class-wise memory bank.
\newblock In {\em Proceedings of the IEEE/CVF International Conference on
  Computer Vision}, pages 8219--8228, 2021.

\bibitem{arnab2021vivit}
Anurag Arnab, Mostafa Dehghani, Georg Heigold, Chen Sun, Mario Lu{\v{c}}i{\'c},
  and Cordelia Schmid.
\newblock Vivit: A video vision transformer.
\newblock In {\em ICCV}, pages 6836--6846, 2021.

\bibitem{detr}
Nicolas Carion, Francisco Massa, Gabriel Synnaeve, Nicolas Usunier, Alexander
  Kirillov, and Sergey Zagoruyko.
\newblock End-to-end object detection with transformers.
\newblock In {\em ECCV}, 2020.

\bibitem{chen2020simple}
Ting Chen, Simon Kornblith, Mohammad Norouzi, and Geoffrey Hinton.
\newblock A simple framework for contrastive learning of visual
  representations.
\newblock In {\em International conference on machine learning}, pages
  1597--1607. PMLR, 2020.

\bibitem{chen2020big}
Ting Chen, Simon Kornblith, Kevin Swersky, Mohammad Norouzi, and Geoffrey~E
  Hinton.
\newblock Big self-supervised models are strong semi-supervised learners.
\newblock {\em Advances in neural information processing systems},
  33:22243--22255, 2020.

\bibitem{cheng2021per}
Bowen Cheng, Alex Schwing, and Alexander Kirillov.
\newblock Per-pixel classification is not all you need for semantic
  segmentation.
\newblock {\em Advances in Neural Information Processing Systems},
  34:17864--17875, 2021.

\bibitem{samtrack}
Yangming Cheng, Liulei Li, Yuanyou Xu, Xiaodi Li, Zongxin Yang, Wenguan Wang,
  and Yi Yang.
\newblock Segment and track anything.
\newblock {\em arXiv preprint arXiv:2305.06558}, 2023.

\bibitem{chitta2021neat}
Kashyap Chitta, Aditya Prakash, and Andreas Geiger.
\newblock Neat: Neural attention fields for end-to-end autonomous driving.
\newblock In {\em Proceedings of the IEEE/CVF International Conference on
  Computer Vision}, pages 15793--15803, 2021.

\bibitem{choi2020pose2mesh}
Hongsuk Choi, Gyeongsik Moon, and Kyoung~Mu Lee.
\newblock Pose2mesh: Graph convolutional network for 3d human pose and mesh
  recovery from a 2d human pose.
\newblock In {\em European Conference on Computer Vision}, pages 769--787.
  Springer, 2020.

\bibitem{3dCrwodNet}
Hongsuk Choi, Gyeongsik Moon, JoonKyu Park, and Kyoung~Mu Lee.
\newblock Learning to estimate robust 3d human mesh from in-the-wild crowded
  scenes.
\newblock In {\em Proceedings of the IEEE/CVF Conference on Computer Vision and
  Pattern Recognition}, pages 1475--1484, 2022.

\bibitem{dosovitskiy2020ViT}
Alexey Dosovitskiy, Lucas Beyer, Alexander Kolesnikov, Dirk Weissenborn,
  Xiaohua Zhai, Thomas Unterthiner, Mostafa Dehghani, Matthias Minderer, Georg
  Heigold, Sylvain Gelly, et~al.
\newblock An image is worth 16x16 words: Transformers for image recognition at
  scale.
\newblock {\em arXiv preprint arXiv:2010.11929}, 2020.

\bibitem{dwivedi2021learning}
Sai~Kumar Dwivedi, Nikos Athanasiou, Muhammed Kocabas, and Michael~J Black.
\newblock Learning to regress bodies from images using differentiable semantic
  rendering.
\newblock In {\em Proceedings of the IEEE/CVF International Conference on
  Computer Vision}, pages 11250--11259, 2021.

\bibitem{fieraru2021remips}
Mihai Fieraru, Mihai Zanfir, Teodor Szente, Eduard Bazavan, Vlad Olaru, and
  Cristian Sminchisescu.
\newblock Remips: Physically consistent 3d reconstruction of multiple
  interacting people under weak supervision.
\newblock {\em Advances in Neural Information Processing Systems},
  34:19385--19397, 2021.

\bibitem{grill2020bootstrap}
Jean-Bastien Grill, Florian Strub, Florent Altch{\'e}, Corentin Tallec, Pierre
  Richemond, Elena Buchatskaya, Carl Doersch, Bernardo Avila~Pires, Zhaohan
  Guo, Mohammad Gheshlaghi~Azar, et~al.
\newblock Bootstrap your own latent-a new approach to self-supervised learning.
\newblock {\em Advances in neural information processing systems},
  33:21271--21284, 2020.

\bibitem{guan2021bilevel}
Shanyan Guan, Jingwei Xu, Yunbo Wang, Bingbing Ni, and Xiaokang Yang.
\newblock Bilevel online adaptation for out-of-domain human mesh
  reconstruction.
\newblock In {\em Proceedings of the IEEE/CVF Conference on Computer Vision and
  Pattern Recognition}, pages 10472--10481, 2021.

\bibitem{gutmann2010noise}
Michael Gutmann and Aapo Hyv{\"a}rinen.
\newblock Noise-contrastive estimation: A new estimation principle for
  unnormalized statistical models.
\newblock In {\em Proceedings of the thirteenth international conference on
  artificial intelligence and statistics}, pages 297--304. JMLR Workshop and
  Conference Proceedings, 2010.

\bibitem{he2020momentum}
Kaiming He, Haoqi Fan, Yuxin Wu, Saining Xie, and Ross Girshick.
\newblock Momentum contrast for unsupervised visual representation learning.
\newblock In {\em Proceedings of the IEEE/CVF conference on computer vision and
  pattern recognition}, pages 9729--9738, 2020.

\bibitem{resnet}
Kaiming He, Xiangyu Zhang, Shaoqing Ren, and Jian Sun.
\newblock Deep residual learning for image recognition.
\newblock In {\em Proceedings of the IEEE conference on computer vision and
  pattern recognition}, pages 770--778, 2016.

\bibitem{hu2021region}
Hanzhe Hu, Jinshi Cui, and Liwei Wang.
\newblock Region-aware contrastive learning for semantic segmentation.
\newblock In {\em Proceedings of the IEEE/CVF International Conference on
  Computer Vision}, pages 16291--16301, 2021.

\bibitem{ionescu_h36m}
Catalin Ionescu, Dragos Papava, Vlad Olaru, and Cristian Sminchisescu.
\newblock {Human3.6M}: Large scale datasets and predictive methods for {3D}
  human sensing in natural environments.
\newblock In {\em TPAMI}, 2014.

\bibitem{jiang2020multiperson}
Wen Jiang, Nikos Kolotouros, Georgios Pavlakos, Xiaowei Zhou, and Kostas
  Daniilidis.
\newblock Coherent reconstruction of multiple humans from a single image.
\newblock In {\em Proceedings of the IEEE/CVF Conference on Computer Vision and
  Pattern Recognition}, pages 5579--5588, 2020.

\bibitem{CMUpanoptic}
Hanbyul Joo, Hao Liu, Lei Tan, Lin Gui, Bart Nabbe, Iain Matthews, Takeo
  Kanade, Shohei Nobuhara, and Yaser Sheikh.
\newblock Panoptic studio: A massively multiview system for social motion
  capture.
\newblock In {\em Proceedings of the IEEE International Conference on Computer
  Vision}, pages 3334--3342, 2015.

\bibitem{joo2021exemplar_EFT}
Hanbyul Joo, Natalia Neverova, and Andrea Vedaldi.
\newblock Exemplar fine-tuning for 3d human model fitting towards in-the-wild
  3d human pose estimation.
\newblock In {\em 2021 International Conference on 3D Vision (3DV)}, pages
  42--52. IEEE, 2021.

\bibitem{kamath2021mdetr}
Aishwarya Kamath, Mannat Singh, Yann LeCun, Gabriel Synnaeve, Ishan Misra, and
  Nicolas Carion.
\newblock Mdetr-modulated detection for end-to-end multi-modal understanding.
\newblock In {\em Proceedings of the IEEE/CVF International Conference on
  Computer Vision}, pages 1780--1790, 2021.

\bibitem{HMR}
Angjoo Kanazawa, Michael~J. Black, David~W. Jacobs, and Jitendra Malik.
\newblock End-to-end recovery of human shape and pose.
\newblock In {\em Computer Vision and Pattern Recognition (CVPR)}, 2018.

\bibitem{kanazawa2019learning}
Angjoo Kanazawa, Jason~Y Zhang, Panna Felsen, and Jitendra Malik.
\newblock Learning 3d human dynamics from video.
\newblock In {\em Proceedings of the IEEE/CVF conference on computer vision and
  pattern recognition}, pages 5614--5623, 2019.

\bibitem{Kendall_2018_CVPR}
Alex Kendall, Yarin Gal, and Roberto Cipolla.
\newblock Multi-task learning using uncertainty to weigh losses for scene
  geometry and semantics.
\newblock In {\em Proceedings of the IEEE Conference on Computer Vision and
  Pattern Recognition (CVPR)}, June 2018.

\bibitem{OCHMR}
Rawal Khirodkar, Shashank Tripathi, and Kris Kitani.
\newblock Occluded human mesh recovery.
\newblock In {\em Proceedings of the IEEE/CVF Conference on Computer Vision and
  Pattern Recognition}, pages 1715--1725, 2022.

\bibitem{kissos2020beyond}
Imry Kissos, Lior Fritz, Matan Goldman, Omer Meir, Eduard Oks, and Mark Kliger.
\newblock Beyond weak perspective for monocular 3d human pose estimation.
\newblock In {\em European Conference on Computer Vision}, pages 541--554.
  Springer, 2020.

\bibitem{kocabas2020vibe}
Muhammed Kocabas, Nikos Athanasiou, and Michael~J Black.
\newblock Vibe: Video inference for human body pose and shape estimation.
\newblock In {\em Proceedings of the IEEE/CVF Conference on Computer Vision and
  Pattern Recognition}, pages 5253--5263, 2020.

\bibitem{PARE}
Muhammed Kocabas, Chun-Hao~P Huang, Otmar Hilliges, and Michael~J Black.
\newblock Pare: Part attention regressor for 3d human body estimation.
\newblock In {\em Proceedings of the IEEE/CVF International Conference on
  Computer Vision}, pages 11127--11137, 2021.

\bibitem{SPIN}
Nikos Kolotouros, Georgios Pavlakos, Michael~J Black, and Kostas Daniilidis.
\newblock Learning to reconstruct 3d human pose and shape via model-fitting in
  the loop.
\newblock In {\em ICCV}, 2019.

\bibitem{GraphCMR}
Nikos Kolotouros, Georgios Pavlakos, and Kostas Daniilidis.
\newblock Convolutional mesh regression for single-image human shape
  reconstruction.
\newblock In {\em Proceedings of the IEEE/CVF Conference on Computer Vision and
  Pattern Recognition}, pages 4501--4510, 2019.

\bibitem{lei2021detecting}
Jie Lei, Tamara~L Berg, and Mohit Bansal.
\newblock Detecting moments and highlights in videos via natural language
  queries.
\newblock {\em Advances in Neural Information Processing Systems},
  34:11846--11858, 2021.

\bibitem{li2018crowdpose}
Jiefeng Li, Can Wang, Hao Zhu, Yihuan Mao, Hao-Shu Fang, and Cewu Lu.
\newblock Crowdpose: Efficient crowded scenes pose estimation and a new
  benchmark.
\newblock {\em arXiv preprint arXiv:1812.00324}, 2018.

\bibitem{li2021hybrik}
Jiefeng Li, Chao Xu, Zhicun Chen, Siyuan Bian, Lixin Yang, and Cewu Lu.
\newblock Hybrik: A hybrid analytical-neural inverse kinematics solution for 3d
  human pose and shape estimation.
\newblock In {\em Proceedings of the IEEE/CVF Conference on Computer Vision and
  Pattern Recognition}, pages 3383--3393, 2021.

\bibitem{li2023seg}
Kexin Li, Zongxin Yang, Lei Chen, Yi Yang, and Jun Xiao.
\newblock Catr: Combinatorial-dependence audio-queried transformer for
  audio-visual video segmentation.
\newblock In {\em Proceedings of the 31th ACM International Conference on
  Multimedia}, 2023.

\bibitem{li2020hero}
Linjie Li, Yen-Chun Chen, Yu Cheng, Zhe Gan, Licheng Yu, and Jingjing Liu.
\newblock Hero: Hierarchical encoder for video+ language omni-representation
  pre-training.
\newblock In {\em Proceedings of the 2020 Conference on Empirical Methods in
  Natural Language Processing (EMNLP)}, pages 2046--2065, 2020.

\bibitem{li2021referring}
Muchen Li and Leonid Sigal.
\newblock Referring transformer: A one-step approach to multi-task visual
  grounding.
\newblock {\em Advances in neural information processing systems},
  34:19652--19664, 2021.

\bibitem{li2022bevformer}
Zhiqi Li, Wenhai Wang, Hongyang Li, Enze Xie, Chonghao Sima, Tong Lu, Yu Qiao,
  and Jifeng Dai.
\newblock Bevformer: Learning bird’s-eye-view representation from
  multi-camera images via spatiotemporal transformers.
\newblock In {\em Computer Vision--ECCV 2022: 17th European Conference, Tel
  Aviv, Israel, October 23--27, 2022, Proceedings, Part IX}, pages 1--18.
  Springer, 2022.

\bibitem{metro_lin}
Kevin Lin, Lijuan Wang, and Zicheng Liu.
\newblock End-to-end human pose and mesh reconstruction with transformers.
\newblock In {\em Proceedings of the IEEE/CVF Conference on Computer Vision and
  Pattern Recognition}, pages 1954--1963, 2021.

\bibitem{mesh_graphormer}
Kevin Lin, Lijuan Wang, and Zicheng Liu.
\newblock Mesh graphormer.
\newblock In {\em Proceedings of the IEEE/CVF International Conference on
  Computer Vision}, pages 12939--12948, 2021.

\bibitem{coco}
Tsung-Yi Lin, Michael Maire, Serge Belongie, James Hays, Pietro Perona, Deva
  Ramanan, Piotr Doll{\'a}r, and C~Lawrence Zitnick.
\newblock Microsoft {COCO}: Common objects in context.
\newblock In {\em ECCV}, 2014.

\bibitem{liu2018coordconv}
Rosanne Liu, Joel Lehman, Piero Molino, Felipe Petroski~Such, Eric Frank, Alex
  Sergeev, and Jason Yosinski.
\newblock An intriguing failing of convolutional neural networks and the
  coordconv solution.
\newblock In {\em Advances in Neural Information Processing Systems}, 2018.

\bibitem{liu2021swin}
Ze Liu, Yutong Lin, Yue Cao, Han Hu, Yixuan Wei, Zheng Zhang, Stephen Lin, and
  Baining Guo.
\newblock Swin transformer: Hierarchical vision transformer using shifted
  windows.
\newblock In {\em Proceedings of the IEEE/CVF international conference on
  computer vision}, pages 10012--10022, 2021.

\bibitem{SMPL}
Matthew Loper, Naureen Mahmood, Javier Romero, Gerard Pons-Moll, and Michael~J
  Black.
\newblock Smpl: A skinned multi-person linear model.
\newblock {\em ACM transactions on graphics (TOG)}, 34(6):1--16, 2015.

\bibitem{AdamW}
Ilya Loshchilov and Frank Hutter.
\newblock Decoupled weight decay regularization.
\newblock In {\em International Conference on Learning Representations}, 2018.

\bibitem{muco}
Dushyant Mehta, Oleksandr Sotnychenko, Franziska Mueller, Weipeng Xu, Srinath
  Sridhar, Gerard Pons-Moll, and Christian Theobalt.
\newblock Single-shot multi-person 3d pose estimation from monocular rgb.
\newblock In {\em 3D Vision (3DV), 2018 Sixth International Conference on}.
  IEEE, sep 2018.

\bibitem{Moon_2022_CVPRW_NeuralAnnot}
Gyeongsik Moon, Hongsuk Choi, and Kyoung~Mu Lee.
\newblock Neuralannot: Neural annotator for 3d human mesh training sets.
\newblock In {\em Computer Vision and Pattern Recognition Workshop (CVPRW)},
  2022.

\bibitem{i2L_meshNet}
Gyeongsik Moon and Kyoung~Mu Lee.
\newblock I2l-meshnet: Image-to-lixel prediction network for accurate 3d human
  pose and mesh estimation from a single rgb image.
\newblock In {\em European Conference on Computer Vision}, pages 752--768.
  Springer, 2020.

\bibitem{oord2018representation}
Aaron van~den Oord, Yazhe Li, and Oriol Vinyals.
\newblock Representation learning with contrastive predictive coding.
\newblock {\em arXiv preprint arXiv:1807.03748}, 2018.

\bibitem{patel2021agora}
Priyanka Patel, Chun-Hao~P Huang, Joachim Tesch, David~T Hoffmann, Shashank
  Tripathi, and Michael~J Black.
\newblock Agora: Avatars in geography optimized for regression analysis.
\newblock In {\em Proceedings of the IEEE/CVF Conference on Computer Vision and
  Pattern Recognition}, pages 13468--13478, 2021.

\bibitem{SMPL-X:2019}
Georgios Pavlakos, Vasileios Choutas, Nima Ghorbani, Timo Bolkart, Ahmed A.~A.
  Osman, Dimitrios Tzionas, and Michael~J. Black.
\newblock Expressive body capture: {3D} hands, face, and body from a single
  image.
\newblock In {\em Proceedings IEEE Conf. on Computer Vision and Pattern
  Recognition (CVPR)}, pages 10975--10985, 2019.

\bibitem{reiher2020sim2real}
Lennart Reiher, Bastian Lampe, and Lutz Eckstein.
\newblock A sim2real deep learning approach for the transformation of images
  from multiple vehicle-mounted cameras to a semantically segmented image in
  bird’s eye view.
\newblock In {\em 2020 IEEE 23rd International Conference on Intelligent
  Transportation Systems (ITSC)}, pages 1--7. IEEE, 2020.

\bibitem{shen2023global}
Xiaolong Shen, Zongxin Yang, Xiaohan Wang, Jianxin Ma, Chang Zhou, and Yi Yang.
\newblock Global-to-local modeling for video-based 3d human pose and shape
  estimation.
\newblock In {\em Proceedings of the IEEE/CVF Conference on Computer Vision and
  Pattern Recognition}, pages 8887--8896, 2023.

\bibitem{ROMP}
Yu Sun, Qian Bao, Wu Liu, Yili Fu, Michael~J Black, and Tao Mei.
\newblock Monocular, one-stage, regression of multiple 3d people.
\newblock In {\em Proceedings of the IEEE/CVF International Conference on
  Computer Vision}, pages 11179--11188, 2021.

\bibitem{BEV}
Yu Sun, Wu Liu, Qian Bao, Yili Fu, Tao Mei, and Michael~J Black.
\newblock Putting people in their place: Monocular regression of 3d people in
  depth.
\newblock In {\em Proceedings of the IEEE/CVF Conference on Computer Vision and
  Pattern Recognition}, pages 13243--13252, 2022.

\bibitem{sun2019human}
Yu Sun, Yun Ye, Wu Liu, Wenpeng Gao, Yili Fu, and Tao Mei.
\newblock Human mesh recovery from monocular images via a skeleton-disentangled
  representation.
\newblock In {\em Proceedings of the IEEE/CVF International Conference on
  Computer Vision}, pages 5349--5358, 2019.

\bibitem{tripathi2020posenet3d}
Shashank Tripathi, Siddhant Ranade, Ambrish Tyagi, and Amit Agrawal.
\newblock Posenet3d: Learning temporally consistent 3d human pose via knowledge
  distillation.
\newblock In {\em 2020 International Conference on 3D Vision (3DV)}, pages
  311--321. IEEE, 2020.

\bibitem{transformer}
Ashish Vaswani, Noam Shazeer, Niki Parmar, Jakob Uszkoreit, Llion Jones,
  Aidan~N. Gomez, Lukasz Kaiser, and Illia Polosukhin.
\newblock Attention is all you need.
\newblock In Isabelle Guyon, Ulrike von Luxburg, Samy Bengio, Hanna~M. Wallach,
  Rob Fergus, S.~V.~N. Vishwanathan, and Roman Garnett, editors, {\em Advances
  in Neural Information Processing Systems 30: Annual Conference on Neural
  Information Processing Systems 2017, December 4-9, 2017, Long Beach, CA,
  {USA}}, pages 5998--6008, 2017.

\bibitem{3dpw}
Timo von Marcard, Roberto Henschel, Michael Black, Bodo Rosenhahn, and Gerard
  Pons-Moll.
\newblock Recovering accurate 3d human pose in the wild using imus and a moving
  camera.
\newblock In {\em European Conference on Computer Vision (ECCV)}, sep 2018.

\bibitem{wang2021exploring}
Wenguan Wang, Tianfei Zhou, Fisher Yu, Jifeng Dai, Ender Konukoglu, and Luc
  Van~Gool.
\newblock Exploring cross-image pixel contrast for semantic segmentation.
\newblock In {\em Proceedings of the IEEE/CVF International Conference on
  Computer Vision}, pages 7303--7313, 2021.

\bibitem{wang2022detr3d}
Yue Wang, Vitor~Campagnolo Guizilini, Tianyuan Zhang, Yilun Wang, Hang Zhao,
  and Justin Solomon.
\newblock Detr3d: 3d object detection from multi-view images via 3d-to-2d
  queries.
\newblock In {\em Conference on Robot Learning}, pages 180--191. PMLR, 2022.

\bibitem{xiao2018simple}
Bin Xiao, Haiping Wu, and Yichen Wei.
\newblock Simple baselines for human pose estimation and tracking.
\newblock In {\em Proceedings of the European conference on computer vision
  (ECCV)}, pages 466--481, 2018.

\bibitem{yang2021projecting}
Weixiang Yang, Qi Li, Wenxi Liu, Yuanlong Yu, Yuexin Ma, Shengfeng He, and Jia
  Pan.
\newblock Projecting your view attentively: Monocular road scene layout
  estimation via cross-view transformation.
\newblock In {\em Proceedings of the IEEE/CVF Conference on Computer Vision and
  Pattern Recognition}, pages 15536--15545, 2021.

\bibitem{mkr}
Yi Yang, Yueting Zhuang, and Yunhe Pan.
\newblock Multiple knowledge representation for big data artificial
  intelligence: framework, applications, and case studies.
\newblock {\em Frontiers of Information Technology \& Electronic Engineering},
  22(12):1551--1558, 2021.

\bibitem{yang2021AOT}
Zongxin Yang, Yunchao Wei, and Yi Yang.
\newblock Associating objects with transformers for video object segmentation.
\newblock {\em Advances in Neural Information Processing Systems},
  34:2491--2502, 2021.

\bibitem{yang2022decoupling}
Zongxin Yang and Yi Yang.
\newblock Decoupling features in hierarchical propagation for video object
  segmentation.
\newblock {\em Advances in Neural Information Processing Systems}, 2022.

\bibitem{yang2021dsc}
Zongxin Yang, Xin Yu, and Yi Yang.
\newblock Dsc-posenet: Learning 6dof object pose estimation via dual-scale
  consistency.
\newblock In {\em Proceedings of the IEEE/CVF Conference on Computer Vision and
  Pattern Recognition}, pages 3907--3916, 2021.

\bibitem{zanfir2018monocular}
Andrei Zanfir, Elisabeta Marinoiu, and Cristian Sminchisescu.
\newblock Monocular 3d pose and shape estimation of multiple people in natural
  scenes-the importance of multiple scene constraints.
\newblock In {\em Proceedings of the IEEE Conference on Computer Vision and
  Pattern Recognition}, pages 2148--2157, 2018.

\bibitem{zanfir2018deep}
Andrei Zanfir, Elisabeta Marinoiu, Mihai Zanfir, Alin-Ionut Popa, and Cristian
  Sminchisescu.
\newblock Deep network for the integrated 3d sensing of multiple people in
  natural images.
\newblock {\em Advances in Neural Information Processing Systems}, 31, 2018.

\bibitem{zhang2021pymaf}
Hongwen Zhang, Yating Tian, Xinchi Zhou, Wanli Ouyang, Yebin Liu, Limin Wang,
  and Zhenan Sun.
\newblock Pymaf: 3d human pose and shape regression with pyramidal mesh
  alignment feedback loop.
\newblock In {\em Proceedings of the IEEE/CVF International Conference on
  Computer Vision}, pages 11446--11456, 2021.

\bibitem{zhang2023rethinking}
Jiangning Zhang, Xiangtai Li, Jian Li, Liang Liu, Zhucun Xue, Boshen Zhang,
  Zhengkai Jiang, Tianxin Huang, Yabiao Wang, and Chengjie Wang.
\newblock Rethinking mobile block for efficient neural models.
\newblock {\em ICCV}, 2023.

\bibitem{zhang2022eatformer}
Jiangning Zhang, Xiangtai Li, Yabiao Wang, Chengjie Wang, Yibo Yang, Yong Liu,
  and Dacheng Tao.
\newblock Eatformer: Improving vision transformer inspired by evolutionary
  algorithm.
\newblock {\em arXiv preprint arXiv:2206.09325}, 2022.

\bibitem{zhang2021analogous}
Jiangning Zhang, Chao Xu, Jian Li, Wenzhou Chen, Yabiao Wang, Ying Tai, Shuo
  Chen, Chengjie Wang, Feiyue Huang, and Yong Liu.
\newblock Analogous to evolutionary algorithm: Designing a unified sequence
  model.
\newblock {\em NeurIPS}, 34:26674--26688, 2021.

\bibitem{CVPE_2020_OOH}
Tianshu Zhang, Buzhen Huang, and Yangang Wang.
\newblock Object-occluded human shape and pose estimation from a single color
  image.
\newblock In {\em Proceedings of the IEEE/CVF conference on computer vision and
  pattern recognition}, pages 7376--7385, 2020.

\bibitem{zhong2021pixel}
Yuanyi Zhong, Bodi Yuan, Hong Wu, Zhiqiang Yuan, Jian Peng, and Yu-Xiong Wang.
\newblock Pixel contrastive-consistent semi-supervised semantic segmentation.
\newblock In {\em Proceedings of the IEEE/CVF International Conference on
  Computer Vision}, pages 7273--7282, 2021.

\bibitem{ddetr}
Xizhou Zhu, Weijie Su, Lewei Lu, Bin Li, Xiaogang Wang, and Jifeng Dai.
\newblock Deformable detr: Deformable transformers for end-to-end object
  detection.
\newblock In {\em ICLR}, 2021.

\end{thebibliography}
}

\end{document}